\documentclass{article}
\usepackage[breaklinks=true,bookmarks=false]{hyperref}
\usepackage{spconf,amsmath,graphicx}
\usepackage{float}
\usepackage{color}
\usepackage{subcaption}
\usepackage{graphicx}
\usepackage{hyperref}

\title{RTSeg: Real-time Semantic Segmentation  Comparative Study}

\makeatletter
 \newcommand{\printfnsymbol}[1]{%
  \textsuperscript{\@fnsymbol{#1}}%
 }
 \makeatother
 
 \name {\thanks{* equally contributing} Mennatullah Siam\printfnsymbol{1}, Mostafa Gamal\printfnsymbol{1}, Moemen Abdel-Razek\printfnsymbol{1}, Senthil Yogamani, Martin Jagersand}
\address{mennatul@ualberta.ca, senthil.yogamani@valeo.com \\ University of Alberta, Valeo Vision Systems, Cairo University}

\begin{document}
%
\maketitle

\begin{abstract}
Semantic segmentation benefits robotics related applications, especially autonomous driving. Most of the research on semantic segmentation only focuses on increasing the accuracy of segmentation models with little attention to computationally efficient solutions. The few work conducted in this direction does not provide principled methods to evaluate the different design choices for segmentation. In this paper, we address this gap by presenting a real-time semantic segmentation benchmarking framework with a decoupled design for feature extraction and decoding methods. The framework is comprised of different network architectures for feature extraction such as VGG16, Resnet18, MobileNet, and ShuffleNet. It is also comprised of multiple meta-architectures for segmentation that define the decoding methodology. These include SkipNet, UNet, and Dilation Frontend. Experimental results are presented on the Cityscapes dataset for urban scenes. The modular design allows novel architectures to emerge, that lead to 143x GFLOPs reduction in comparison to SegNet. This benchmarking framework is publicly available at \footnote{https://github.com/MSiam/TFSegmentation}.

\end{abstract}
\section{Introduction}
Semantic segmentation has made progress in the recent years with deep learning. The first prominent work in this field was fully convolutional networks(FCNs) \cite{long2015fully}. FCN was proposed as an end-to-end method to learn pixel-wise classification, where transposed convolution was used for upsampling. Skip architecture was used to refine the segmentation output, that utilized higher resolution feature maps. That method paved the road to subsequent advances in the segmentation accuracy. Multi-scale approaches \cite{chen2016deeplab,yu2015multi}, structured models \cite{lin2016exploring,zheng2015conditional}, and spatio-temporal architectures \cite{shelhamer2016clockwork} introduced different directions for improving accuracy. All of the above approaches focused on accuracy and robustness of segmentation. Well known benchmarks and datasets for semantic segmentation such as Pascal \cite{everingham2010pascal},  NYU RGBD \cite{silberman2012indoor}, Cityscapes \cite{cordts2016cityscapes}, and Mapillary \cite{neuhold2017mapillary} boosted the competition toward improving accuracy.

\begin{figure}
\centering
\includegraphics[scale= 0.45]{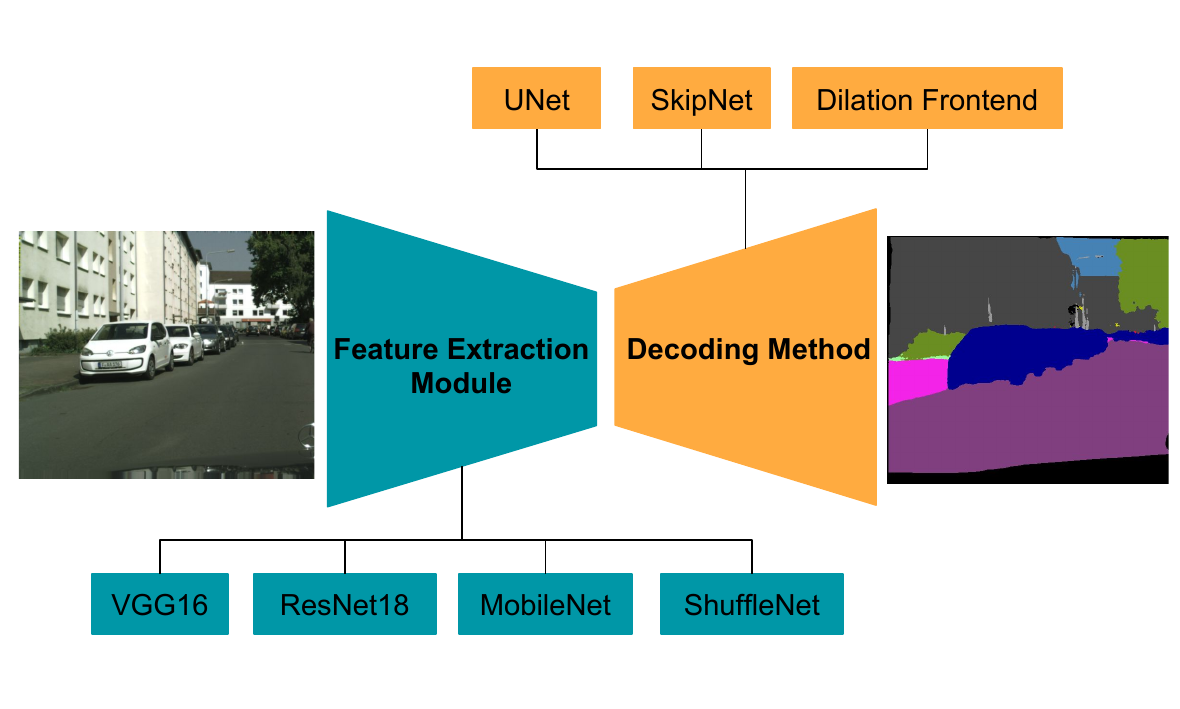}
\caption{Overview of the different components in the framework with the decoupling of feature extraction module and decoding method. }
\label{fig:overview}
\end{figure}

\begin{figure*}[ht!]
\centering
\begin{subfigure}{.5\textwidth}
    \includegraphics[scale= 0.13]{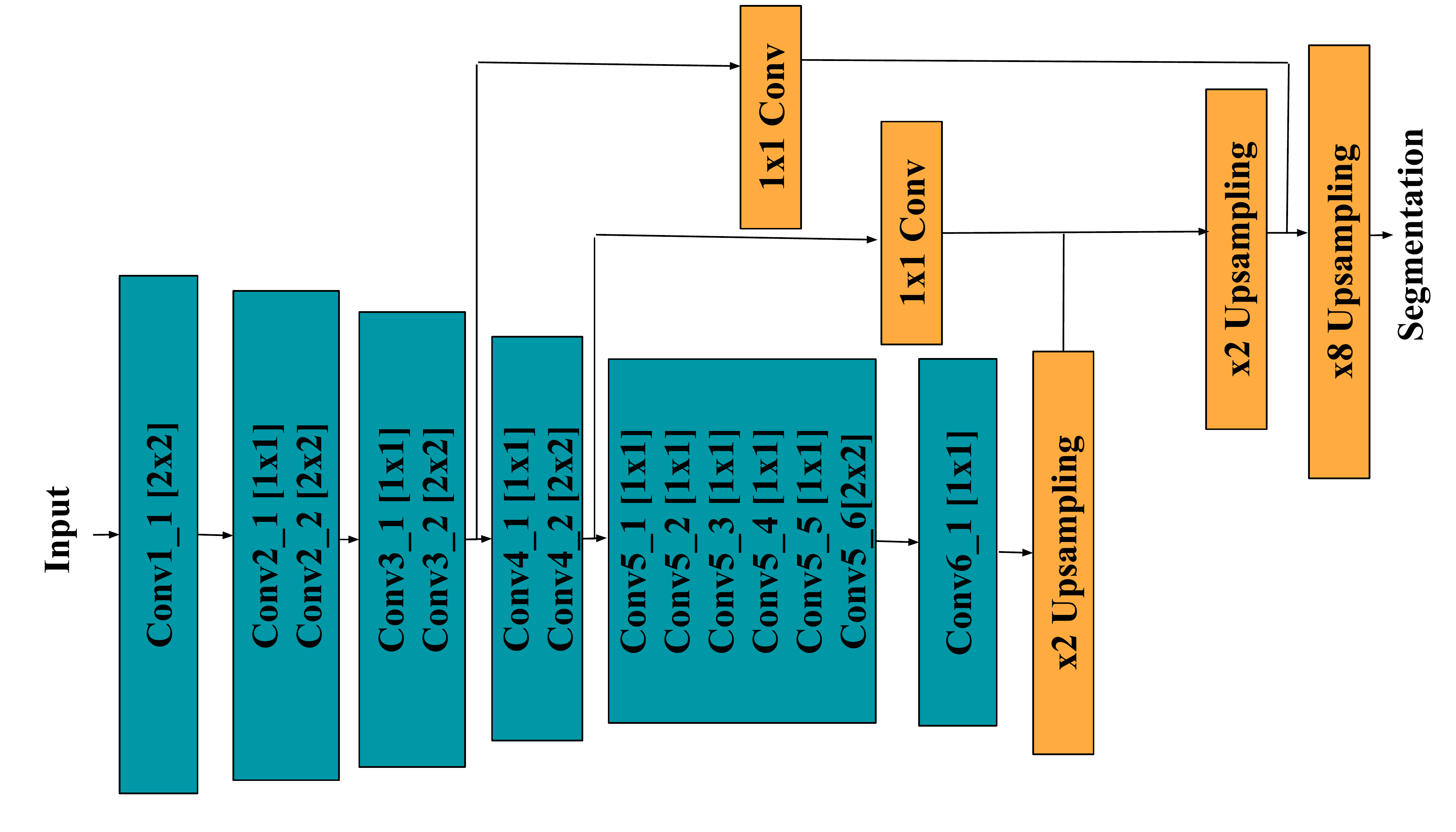}
    \caption{}
    \end{subfigure}%
\begin{subfigure}{.5\textwidth}
    \includegraphics[scale= 0.13]{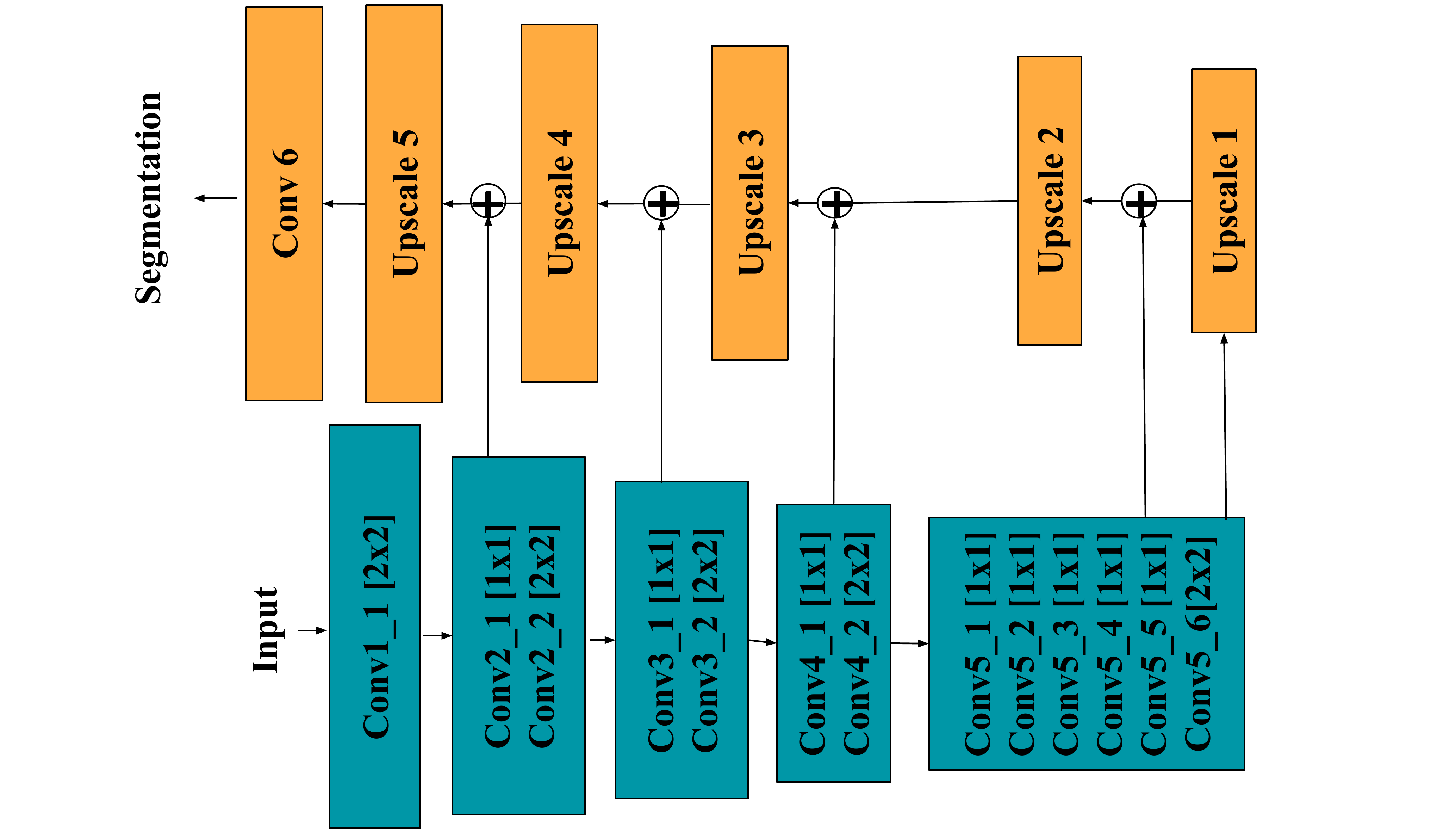}
    \caption{}
\end{subfigure}
\caption{Different Meta Architectures using MobileNet as the feature extraction network. a) SkipNet. b) UNet. }
\label{fig:architecture}
\end{figure*}

However, little attention is given to the computational efficiency of these networks. Although, when it comes to applications such as autonomous driving this would have tremendous impact. There exists few work that tries to address the segmentation networks efficiency such as \cite{zhao2017icnet,paszke2016enet}. The survey on semantic segmentation \cite{garcia2017review} presented a comparative study between different segmentation architectures including ENet \cite{paszke2016enet}. Yet, there is no principled comparison of different networks and meta-architectures. These previous studies compared different networks as a whole, without comparing the effect of different modules. That does not enable researchers and practitioners to pick the best suited design choices for the required task.

In this paper we propose the first framework toward benchmarking real-time architectures in segmentation. Our main contributions are: (1) we provide a modular decoupling of the segmentation architecture into feature extraction and decoding method which is termed as meta-architecture as shown in Figure \ref{fig:overview}. The separation helps in understanding the impact of different parts of the network on real-time performance. (2) A detailed ablation study with highlighting the trade-off between accuracy and computational efficiency is presented. (3) The modular design of our framework allowed the emergence of two novel segmentation architectures using MobileNet \cite{howard2017mobilenets} and ShuffleNet \cite{zhang2017shufflenet} with multiple decoding methods. ShuffleNet lead to 143x GFLOPs reduction in comparison to SegNet. Our framework is built on top of Tensorflow and is publicly available. Although our framework delivers less operations we have not been able to deliver higher inference speed.

\section{Benchmarking Framework}
\subsection{Meta-Architectures}
Three meta-architectures are integrated in our benchmarking software: (1) SkipNet meta-architecture\cite{long2015fully}. (2) U-Net meta-architecture\cite{ronneberger2015u}. (3) Dilation Frontend meta-architecture\cite{yu2015multi}.
The meta-architectures for semantic segmentation identify the decoding method for in the network upsampling.  All of the network architectures share the same down-sampling factor of 32. The downsampling is achieved either by utilizing pooling layers, or strides in the convolutional layers. This ensures that different meta architectures have a unified down-sampling factor to assess the effect of the decoding method only. 
\begin{table*}[ht!]
\centering
\caption{Comparison of different encoders and decoding methods in accuracy on cityscapes validation set. The modular decoupled design in RTSeg enabled such comparison. Coarse indicates whether the network was pre-trained on the coarse annotation or not.}
\label{table:ablation}
\begin{tabular}{|l|l|l|l|l|l|l|l|l|l|l|l|l|}
\hline
Decoder & Encoder & Coarse & mIoU & Road & Sidewalk & Building & Sign & Sky & Person & Car & Bicycle & Truck\\ \hline
SkipNet & MobileNet & No & 61.3 & \textbf{95.9} & 73.6 & \textbf{86.9} & 57.6 & 91.2 & 66.4 & \textbf{89.0} & 63.6 & \textbf{45.9}\\ \hline
SkipNet & ShuffleNet & No & 55.5 & 94.8 & 68.6 & 83.9 & 50.5 & 88.6 & 60.8 & 86.5 & 58.8 & 29.6\\ \hline
UNet & ResNet18 & No & 57.9 & 95.8 & 73.2 & 85.8 & 57.5 & 91.0 & 66.0 & 88.6 & 63.2 & 31.4 \\ \hline
UNet & MobileNet & No & 61.0 & 95.2 & 71.3 & 86.8 & \textbf{60.9} & \textbf{92.8} & \textbf{68.1} & 88.8 & \textbf{65.0} & 41.3 \\ \hline
UNet & ShuffleNet & No & 57.0 & 95.1 & 69.5 & 83.7 & 54.3 & 89.0 & 61.7 & 87.8 & 59.9 & 35.5\\ \hline
Dilation & MobileNet & No & 57.8 & 95.6 & 72.3 & 85.9 & 57.0 & 91.4 & 64.9 & 87.8 & 62.8 & 26.3 \\ \hline
Dilation & ShuffleNet & No & 53.9 & 95.2 & 68.5 & 84.1 & 57.3 & 90.3 & 62.9 & 86.6 & 60.2 & 23.3\\ \hline
SkipNet & MobileNet & Yes & \textbf{62.4} & 95.4 & \textbf{73.9} & 86.6 & 57.4 & 91.1 & 65.7 & 88.4 & 63.3 & 45.3 \\ \hline
SkipNet & ShuffleNet & Yes & 59.3 & 94.6 & 70.5 & 85.5 & 54.9 & 90.8 & 60.2 & 87.5 & 58.8 & 45.4\\ \hline
\end{tabular}
\end{table*}

\textbf{SkipNet} architecture denotes a similar architecture to FCN8s \cite{long2015fully}. The main idea of the skip architecture is to benefit from feature maps from higher resolution to improve the output segmentation. SkipNet applies transposed convolution on heatmaps in the label space instead of performing it on feature space. This entails a more computationally efficient decoding method than others. Feature extraction networks have the same downsampling factor of 32, so they follow the 8 stride version of skip architecture. Higher resolution feature maps are followed by 1x1 convolution to map from feature space to label space that produces heatmaps corresponding to each class. The final heatmap with downsampling factor of 32 is followed by transposed convolution with stride 2. Elementwise addition between this upsampled heatmaps and the higher resolution heatmaps is performed. Finally, the output heat maps are followed by a transposed convolution for up-sampling with stride 8. Figure \ref{fig:architecture}(a) shows the SkipNet architecture utilizing a MobileMet encoder.

\textbf{U-Net} architecture denotes the method of decoding that up-samples features using transposed convolution corresponding to each downsampling stage. The up-sampled features are fused with the corresponding features maps from the encoder with the same resolution. The stage-wise upsampling provides higher accuracy than one shot 8x upsampling. The current fusion method used in the framework is element-wise addition. Concatenation as a fusion method can provide better accuracy, as it enables the network to learn the weighted fusion of features. Nonetheless, it increases the computational cost, as it is directly affected by the number of channels. The upsampled features are then followed by 1x1 convolution to output the final pixel-wise classification. Figure \ref{fig:architecture}(b) shows the UNet architecture using MobileNet as a feature extraction network.

\textbf{Dilation Frontend} architecture utilizes dilated convolution instead of downsampling the feature maps. Dilated convolution enables the network to maintain an adequate receptive field, but without degrading the resolution from pooling or strided convolution. However, a side-effect of this method is that computational cost increases, since the operations are performed on larger resolution feature maps.  The encoder network is modified to incorporate a downsampling factor of 8 instead of 32. The decrease of the downsampling is performed by either removing pooling layers or converting stride 2 convolution to stride 1. The pooling or strided convolutions are then replaced with two dilated convolutions\cite{yu2015multi} with dilation factor 2 and 4 respectively. 

\subsection{Feature Extraction Architectures}
In order to achieve real-time performance multiple network architectures are integrated in the benchmarking framework. The framework includes four state of the art real-time network architectures for feature extraction. These are: (1) VGG16\cite{simonyan2014very}. (2) ResNet18\cite{he2016deep}. (3) MobileNet\cite{howard2017mobilenets}. (4) ShuffleNet \cite{zhang2017shufflenet}. The reason for using \textbf{VGG16} is to act as a baseline method to compare against as it was used in \cite{long2015fully}. The other architectures have been used in real-time systems for detection and classification. \textbf{ResNet18} incorporates the usage of residual blocks that directs the network toward learning the residual representation on identity mapping.

\textbf{MobileNet} network architecture is based on depthwise separable convolution. It is considered the extreme case of the inception module, where separate spatial convolution for each channel is applied denoted as depthwise convolutions. Then 1x1 convolution with all the channels to merge the output denoted as pointwise convolutions is used. The separation in depthwise and pointwise convolution improve the computational efficiency on one hand. On the other hand it improves the accuracy as the cross channel and spatial correlations mapping are learned separately. 

\textbf{ShuffleNet} encoder is based on grouped convolution that is a generalization of depthwise separable convolution. It uses channel shuffling to ensure the connectivity between input and output channels.  This eliminates connectivity restrictions posed by the grouped convolutions.
\label{sec:method}

\section{Experiments}
In this section experimental setup, detailed ablation study and results in comparison to the state of the art are reported.

\subsection{Experimental Setup}
Through all of our experiments, weighted cross entropy loss from \cite{paszke2016enet} is used, to overcome the class imbalance. Adam optimizer \cite{kingma2014adam} learning rate is set to 1$e^{-4}$. Batch normalization \cite{ioffe2015batch} is incorporated. L2 regularization with weight decay rate of 5$e^{-4}$ is utilized to avoid over-fitting. The feature extractor part of the network is initialized with the pre-trained corresponding encoder trained on Imagenet. A width multiplier of 1 for MobileNet to include all the feature channels is performed through all the experiments. The number of groups used in ShuffleNet is 3. Based on previous \cite{zhang2017shufflenet} results on classification and detection three groups provided adequate accuracy. 

Results are reported on Cityscapes dataset \cite{cordts2016cityscapes} which contains 5000 images with fine annotation, with 20 classes including the ignored class. Another section of the dataset contains coarse annotations with 20,000 labeled images. These are used in the case of Coarse pre-training that improves the results of the segmentation. Experiments are conducted on images with resolution of 512x1024.

\begin{table}[ht!]
\centering
\caption{Comparison of the most promising models in our benchmarking framework in terms of GFLOPs and frames per second, this is computed on image resolution 512x1024.}
\label{table:performance}
\begin{tabular}{|l|l|}
\hline
Model & GFLOPs \\ \hline
SkipNet-MobileNet & 13.8 \\ \hline
UNet-MobileNet & 55.9 \\ \hline
\end{tabular}
\end{table}

\begin{table*}[ht!]
\centering
\caption{Comparison of some of the models from our benchmarking framework with the state of the art segmentation networks on cityscapes test set. GFLOPs is computed on image resolution 360x640.}
\label{table:quant_city}
\begin{tabular}{|l|l|l|l|l|l|}
\hline
Model & GFLOPs & Class IoU & Class iIoU & Category IoU & Category iIoU \\ \hline
SegNet\cite{badrinarayanan2015segnet} & 286.03 & 56.1 & 34.2 & 79.8 & 66.4 \\ \hline
ENet\cite{paszke2016enet} &  3.83 & 58.3 & 24.4 & 80.4 & 64.0 \\ \hline
DeepLab\cite{chen2016deeplab} & - & \textbf{70.4} & \textbf{42.6} & \textbf{86.4} & 67.7 \\ \hline
SkipNet-VGG16\cite{long2015fully} & - & 65.3 & 41.7 & 85.7 & \textbf{70.1} \\ \hline
SkipNet-ShuffleNet & \textbf{2.0} & 58.3 & 32.4 & 80.2 & 62.2 \\ \hline
SkipNet-MobileNet & 6.2 & 61.5 & 35.2 & 82.0 & 63.0 \\ \hline
\end{tabular}
\end{table*}

\begin{figure*}[ht!]
\centering
\begin{subfigure}{0.38\textwidth}
    \includegraphics[scale= 0.18]{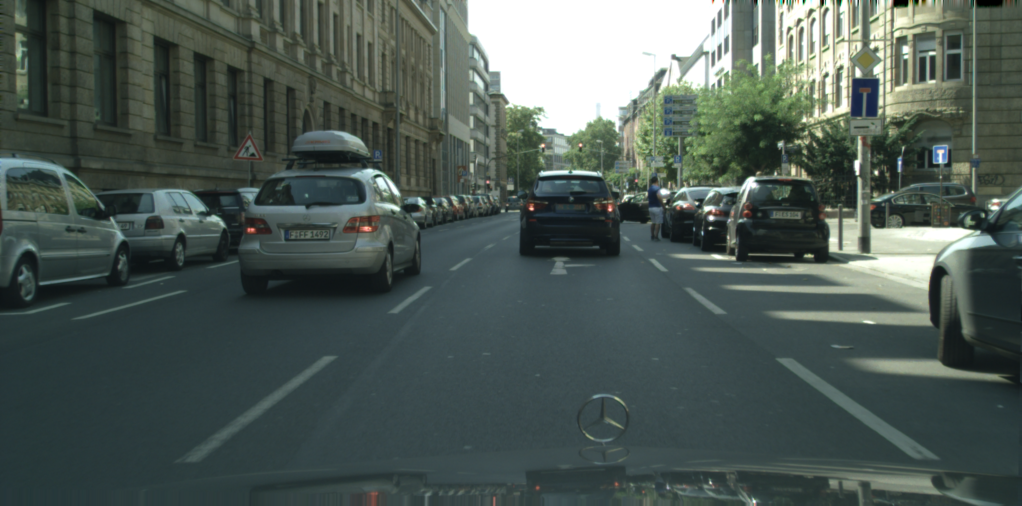}
    \caption{}
\end{subfigure}%
\begin{subfigure}{0.38\textwidth}
    \includegraphics[scale= 0.18]{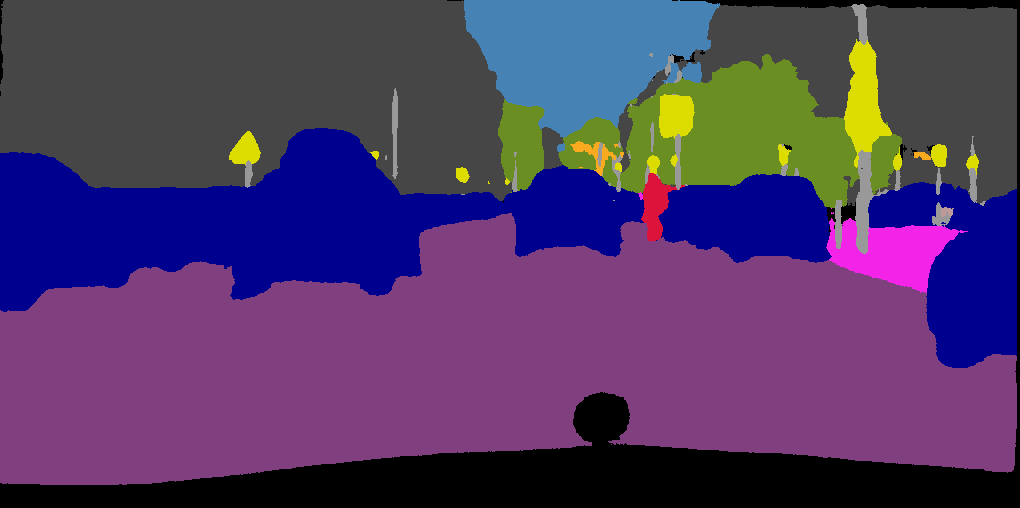}
    \caption{}
\end{subfigure}

\begin{subfigure}{0.38\textwidth}
    \includegraphics[scale= 0.18]{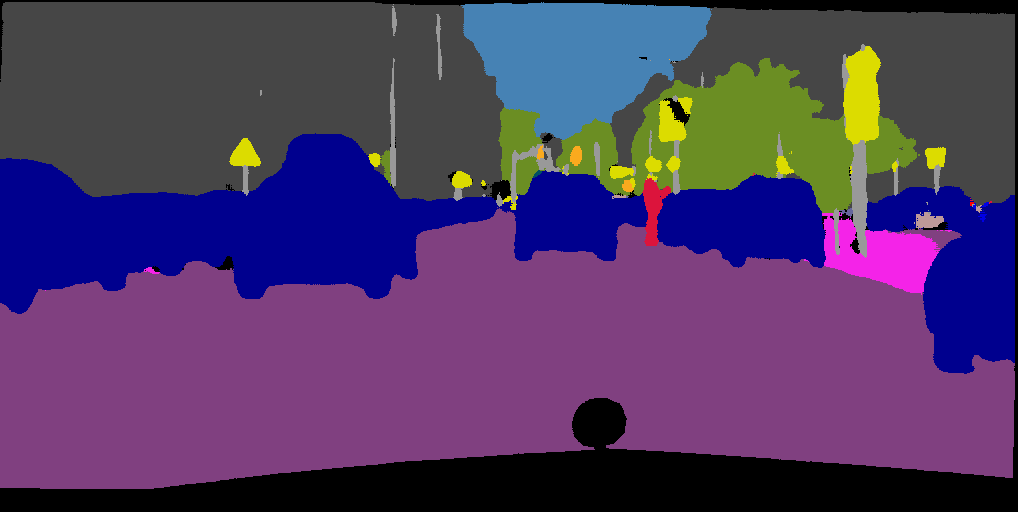}
    \caption{}
\end{subfigure}%
\begin{subfigure}{0.38\textwidth}
    \includegraphics[scale= 0.18]{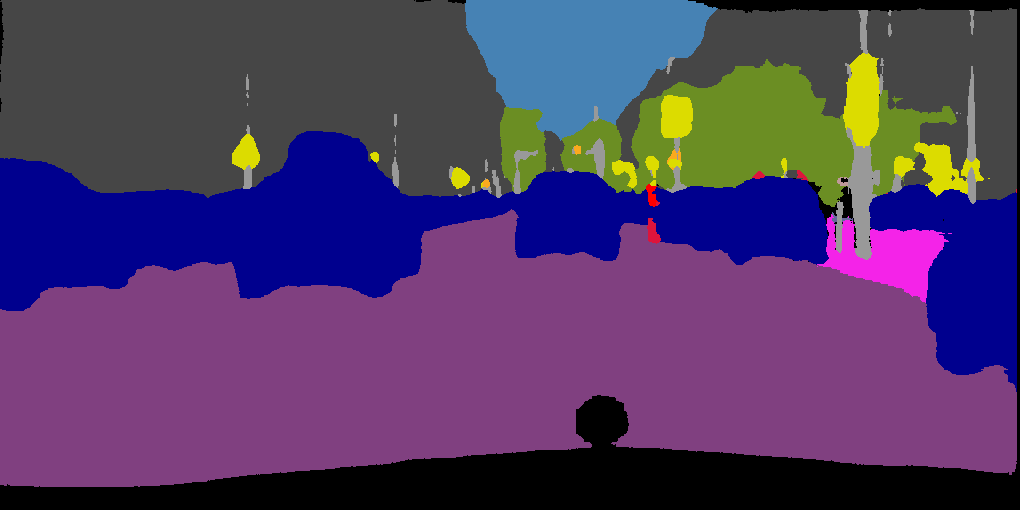}
    \caption{}
\end{subfigure}
\caption{Qualitative Results on CityScapes. (a) Original Image. (b) SkipNet-MobileNet pretrained with Coarse Annotations. (c) UNet-Resnet18. (d) SkipNet-ShuffleNet pretrained with Coarse Annotations.}
\label{fig:qual_city}
\end{figure*}

\subsection{Semantic Segmentation Results}
Semantic segmentation is evaluated using mean intersection over union (mIoU), per-class IoU, and per-category IoU. Table\ref{table:ablation} shows the results for the ablation study on different encoders-decoders with mIoU and GFLOPs to demonstrate the accuracy and computations trade-off. The main insight gained from our experiments is that, UNet decoding method provides more accurate segmentation results than Dilation Frontend. This is mainly due to the transposed convolution by 8x in the end of the Dilation Frontend, unlike the UNet stage-wise upsampling method. The SkipNet architecture provides on par results with UNet decoding method. In some architectures such as SkipNet-ShuffleNet it is less accurate than UNet counter part by 1.5\%. 

The UNet method of incrementally upsampling with-in the network provides the best in terms of accuracy. However, Table \ref{table:performance} clearly shows that SkipNet architecture is more computationally efficient with 4x reduction in GFLOPs. This is explained by the fact that transposed convolutions in UNet are applied in the feature space unlike in SkipNet that are applied in label space. Table \ref{table:ablation} shows that Coarse pre-training improves the overall mIoU with 1-4\%. The underrepresented classes are the ones that often benefit from pre-training.

Experimental results on the cityscapes test set are shown in Table \ref{table:quant_city}. Although, DeepLab provides best results in terms of accuracy, it is not computationally efficient. ENet \cite{paszke2016enet} is compared to SkipNet-ShuffleNet and SkipNet-MobileNet in terms of accuracy and GFLOPs. SkipNet-ShuffleNet outperforms ENet in terms of GFLOPs, yet it maintains on par mIoU. However, we have not been able to outperform ENet in terms of inference speed. Both SkipNet-ShuffleNet and SkipNet-MobileNet outperform SegNet \cite{badrinarayanan2015segnet} in terms of computational cost and accuracy with reduction up to 143x in GFLOPs. Figure \ref{fig:qual_city} shows qualitative results for different encoders including MobileNet, ShuffleNet and ResNet18. It shows that MobileNet provides more accurate segmentation results than the later two. SkipNet-MobileNet is able to correctly segment the pedestrian and the signs on the right unlike the others.

\label{sec:exps}

\section{Conclusion}
In this paper we present the first principled approach for benchmarking real-time segmentation networks. The decoupled design of the framework separates modules for better quantitative comparison. The first module is comprised of the feature extraction network architecture, the second is the meta-architecture that provides the decoding method. Three different meta-architectures are included in our framework, including Skip architecture, UNet, and Dilation Frontend. Different network architectures for feature extraction are included, which are ShuffleNet, MobileNet, VGG16, and ResNet-18. Our benchmarking framework provides researchers and practitioners with a mean to evaluate design choices for their tasks.
\label{sec:conc}

\bibliographystyle{IEEEbib}
\bibliography{refs}

\end{document}